\documentclass[runningheads,a4paper]{llncs}
\mainmatter
\usepackage{graphicx}
\usepackage{amsmath,amssymb} % define this before the line numbering.
\usepackage{color}
\usepackage{graphicx} % more modern
\usepackage{times}
\usepackage{helvet}
\usepackage{courier}
\usepackage{amssymb,amsfonts,verbatim}
\usepackage{amsfonts}

\usepackage{algorithm}
\usepackage{amsmath}

\usepackage{multirow}
\usepackage{url}

\def\C{{\bf C}}

\def\X{{\bf X}}

\def\0{{\bf 0}}
\def\1{{\bf 1}}

\graphicspath{{figure/}}

\title{Robust and Efficient Subspace Segmentation via Least Squares Regression} % Replace with your title

\author{Can-Yi Lu$^{\dagger\ddagger}$\and Hai Min$^{\dagger}$ \and Zhong-Qiu Zhao$^\S$ \and Lin Zhu$^{\dagger}$   \and De-Shuang Huang$^\star$ \and Shuicheng Yan$^\ddagger$}
\institute{$^\dagger$ Department of Automation, University of Science and Technology of China, Hefei, China \\
$^\ddagger$ Department of Electrical and Computer Engineering, National University of Singapore \\
$^\S$ School of Computer and Information, Hefei University of Technology, Hefei, China \\
$^\star$ School of Electronics and Information Engineering, Tongji University, Shanghai, China\\
{\tt\small \{canyilu, minhai361, zhongqiuzhao\}@gmail.com}\\
{\tt\small zhomlynn@yahoo.com.cn} ~~
{\tt\small dshuang@tongji.edu.cn} ~~
{\tt\small eleyans@nus.edu.sg}
}

\begin{document}

\maketitle

\begin{abstract}
This paper studies the subspace segmentation problem which aims to segment data drawn from a union of multiple linear subspaces. Recent works by using sparse representation, low rank representation and their extensions attract much attention. If the subspaces from which the data drawn are independent or orthogonal, they are able to obtain a block diagonal affinity matrix, which usually leads to a correct segmentation. The main differences among them are their objective functions. We theoretically show that if the objective function satisfies some conditions, and the data are sufficiently drawn from independent subspaces, the obtained affinity matrix is always block diagonal. Furthermore, the data sampling can be insufficient if the subspaces are orthogonal. Some existing methods are all special cases. Then we present the Least Squares Regression (LSR) method for subspace segmentation. It takes advantage of data correlation, which is common in real data. LSR encourages a grouping effect which tends to group highly correlated data together. Experimental results on the Hopkins 155 database and Extended Yale Database B show that our method significantly outperforms state-of-the-art methods. Beyond segmentation accuracy, all experiments demonstrate that LSR is much more efficient.
\end{abstract}

\section{Introduction}
Subspace segmentation is an important clustering problem which attracts much attention in recent years. It arises in numerous applications in machine learning and computer vision literature, e.g. image representation \cite{Hong06}, clustering \cite{facecluster} and motion segmentation \cite{motionseg1} \cite{Motion_eccv06}. Given a set of data drawn from a union of subspaces, the goal of subspace segmentation is to segment (cluster or group) data into clusters with each cluster corresponding to a subspace. This problem is formally defined as follow:

\begin{definition}
(Subspace Segmentation) Given a set of data vectors $X=[X_1,\cdots,\\X_k]=[x_1,\cdots,x_n]\in\mathbb{R}^{d\times n}$ drawn from a union of $k$ subspaces $\{\mathcal{S}_i\}_{i=1}^k$. Let $X_i$ ba a collection of $n_i$ data vectors drawn from the subspace $\mathcal{S}_i$, $n=\sum_{i=1}^kn_i$. The task is to segment the data according to the underlying subspaces they are drawn from.
\end{definition}

\subsection{Prior Works}
During the past two decades, a number of subspace segmentation methods have been proposed. According to their mechanisms of representing the subspaces, existing works can be roughly divided into four categories: algebraic methods \cite{Costeria98} \cite{GPCA}, iterative methods \cite{kplane} \cite{Kflats}, statistical methods \cite{Tipping99} \cite{Mayi07} and spectral clustering-based methods \cite{SSC} \cite{LRR} \cite{MSR} \cite{SSQP}. A review of these methods can be found in \cite{tutorialSub}. In this work, we review four most recent and related methods: Sparse Subspace Clustering (SSC) \cite{SSC} \cite{disjointsub}, Low-Rank Representation (LRR) \cite{LRR} \cite{LRRpami} and their extensions \cite{MSR} \cite{SSQP}.

SSC and LRR are two spectral clustering-based methods. A main challenge in applying spectral clustering to subspace segmentation is to define a "good" affinity matrix (or graph) $Z\in\mathbb{R}^{n\times n}$. Each entry $Z_{ij}$ measures the similarity between data points $x_i$ and $x_j$. Ideally, the affinity matrix should be block diagonal, the between-cluster affinities are all zeros. Then it is easy to segment data on such well defined graph by spectral clustering. Typical choice for the measure of similarity is $Z_{ij}=\exp(-||x_i-x_j||/\sigma)$, $\sigma>0$. However, this method is not able to characterize the structure of data from subspaces, and the affinity matrix is also not block diagonal. SSC and LRR provide a new way to construct the affinity matrix. They express each data point $x_i$ as a linear combination of all other data $x_i=\sum_{j\neq i}Z_{ij}x_j$, and use the representational coefficient $(|Z_{ij}|+|Z_{ji}|)/2$ to measure the similarity between $x_i$ and $x_j$. The difference between SSC and LRR lies in the regularization on $Z$: SSC enforces $Z$ to be sparse while LRR encourages $Z$ to be of low-rank.
	
Motivated by the fact that an ideal affinity matrix is block diagonal or sparse, SSC solves the following sparse representation problem:
\begin{equation}
\label{Eq_SSCl0}
\min ||Z||_0 \ \ \text{s.t.} \ X=XZ, \ \text{diag}(Z)=0,
\end{equation}
where $||Z||_0$ is the $\ell^0$-norm of $Z$, \emph{i.e.} the number of nonzero elements. But such optimization problem is non-convex and NP-hard. Under some condition \cite{L0L1equ}, it is equal to the following $\ell^1$-minimization problem:
\begin{equation}
\label{Eq_SSCl1}
\min ||Z||_1 \ \ \text{s.t.} \ X=XZ, \ \text{diag}(Z)=0,
\end{equation}
where $||Z||_1$ denotes the $\ell^1$-norm of $Z$, $||Z||_1=\sum_{i=1}^n\sum_{j=1}^n|Z_{ij}|$. It has been shown that when the subspaces are independent \footnote{A collection of $k$ subspaces $\{\mathcal{S}_i\}_{i=1}^k$ are independent if and only if $\sum_{i=1}^k\mathcal{S}_i=\oplus_{i=1}^k\mathcal{S}_i$.}, the solution to problem (\ref{Eq_SSCl1}) is block diagonal. But this solution does not guarantee to obtain a correct segmentation, since it may be "too sparse", which divides the within-cluster data into different groups. If there is a group of data points among which the pairwise correlations are very high, then sparse representation tends to select only one, at random \cite{Elasticnet}. Thus, SSC is not able to capture the correlation structure of data from the same subspace. Another drawback of SSC is that it is not efficient for solving the $\ell^1$-minimization problem (\ref{Eq_SSCl1}) for each data point.

LRR aims to take the correlation structure of data into account, it finds a low rank representation instead of a sparse representation. In the case of noise free data drawn from linear subspaces, the original LRR solves the following rank minimization problem:
\begin{equation}
\label{Eq_LRRrank}
\min \text{rank}(Z) \ \ \text{s.t.} \ X=XZ,
\end{equation}
where rank$(Z)$ denotes the rank of $Z$. The rank minimization problem is also NP-hard, a common surrogate of rank function is the nuclear norm:
\begin{equation}
\label{Eq_LRR}
\min ||Z||_* \ \ \text{s.t.} \ X=XZ,
\end{equation}
where $||Z||_*$ is the nuclear norm of $Z$, \emph{i.e.} the sum of all the singular values of $Z$. It can be shown that when the subspaces are independent, the solution to problem (\ref{Eq_LRR}) is unique and block diagonal. The extended LRR model for data with noise is efficient and effective for subspace segmentation. However, is it really the low rank property of $Z$ that makes LRR powerful for subspace segmentation? The intuition is not clear. It is not necessary to require a block diagonal matrix to be of low rank. The following is a special example based on the problem (\ref{Eq_LRRrank}):

\emph{Example 1:} Let $X_1=\begin{bmatrix}1 & 2 \\0 & 0\end{bmatrix}$ and $X_2=\begin{bmatrix}0 & 0 \\1 & 2\end{bmatrix}$ are some data points drawn from subspaces $\mathcal{S}_1$ and $\mathcal{S}_2$, respectively. We assume $\text{dim}(\mathcal{S}_1)=\text{dim}(\mathcal{S}_2)=1$, thus $\mathcal{S}_1$ and $\mathcal{S}_2$ are orthogonal \footnote{The orthogonal subspaces must be independent, but not vice versa.}, and $\text{rank}(X_1)=\text{rank}(X_2)=1$. Let $X=[X_1,X_2]$. The solutions to the problem (\ref{Eq_LRRrank}) are not unique and one of which is
\begin{equation}
\label{Eq_example}
Z^*=\begin{bmatrix} 0.5 & 1 & 1 & 2\\0.25 & 0.5 & -0.5 & -1\\ 1 & 2 & 0.5 & 1 \\ -0.5 & -1 & 0.25 & 0.5\end{bmatrix}
\end{equation}
The above example shows that even when the subspaces are orthogonal, the solution to the original LRR model (\ref{Eq_LRRrank}) does not guarantee to be block diagonal, and we are not able to get the true segmentation from (\ref{Eq_example}). But we will theoretically show that it is easy to get a block diagonal solution for independent subspaces segmentation and even the data can be insufficient if the subspaces are orthogonal. Though LRR performs well by solving the nuclear norm minimization problem (\ref{Eq_LRR}) which also leads to a solution with low rank, its superiority may have nothing to do with the rank of the solution. We may regard the nuclear norm (\ref{Eq_LRR}) as a new criterion but not a simple surrogate of the rank function, at least for the subspace segmentation problem. The motivation of LRR comes from RPCA \cite{RPCA}, but one should notice that their physical meanings are very different. RPCA aims to find a low-rank recovery of the observation $X$ while LRR focuses on the matrix of representational coefficient.

Luo \emph{et al.} proposes the Multi-Subspace Representation (MSR) \cite{MSR} which combines both criteria of SSC and LRR by solving the following problem:
\begin{equation}
\label{Eq_MSR}
\min ||Z||_*+\delta||Z||_1 \ \ \text{s.t.} \ X=XZ, \text{diag}(Z)=0.
\end{equation}
Another method, namely Subspace Segmentation via Quadratic Programming (SSQP) \cite{SSQP} is proposed by solving the following problem:
\begin{equation}
\label{Eq_SSQP}
\min ||XZ-X||_F^2+\lambda||Z^TZ||_1 \ \ \text{s.t.} \ Z\geq0, \text{diag}(Z)=0.
\end{equation}

The main differences among the above four methods are their objective functions. All these methods have a similar property: the solutions obtained by SSC, LRR and MSR are block diagonal when the subspaces are independent, and SSQP requires an orthogonal subspaces assumption. Now a question is raised: what kind of objective function and data assumption guarantee a block diagonal solution? Furthermore, which will be a better choice? We try to answer these questions which are summarized in the contributions of this paper.

\subsection{Paper Contributions}
In this work, we first theoretically show that if the objective function satisfies certain conditions, we are able to get a block diagonal solution based on the independent subspaces assumption. The above four criteria all satisfy these conditions and they are all special cases. If the sampling data are not sufficient, we further require the subspaces to be orthogonal. Second, as a new special case, we present the Least Squares Regression (LSR) model for subspace segmentation. The grouping effect of LSR help it group highly correlated data together. We further show that LSR is robust to a bounded disturbance, and is very efficient due to a closed form solution. At last, we experimentally show that LSR is more effective and efficient than SSC and LRR on the Hopkins 155 database and Extended Yale Database B.

\section{Theoretical Analysis}
\label{Se_TheoAna}
For the ease of exploration, we assume the data are noise free in this section. SSC, LRR, MSR and SSQP use different criteria to control the within-cluster and between-cluster affinities. Under certain data assumption (independent or orthogonal), a block diagonal solution is obtained for true segmentation. A natural question is raised: what kind of criterion owns such property? We first consider a simple situation by using a basis of subspaces as dictionary.
\begin{theorem}
\label{Thm_Blocksys}
Assume the subspaces $\{\mathcal{S}_i\}_{i=1}^k$ are independent, $B_i$ is a matrix whose columns consist of a basis of the subspace $\mathcal{S}_i$, $B=[B_1,\cdots,B_k]$, and $X_i$ is a matrix whose columns consist of some vectors from $\mathcal{S}_i$, $X=[X_1,\cdots,X_k]$. The solution $Z^*$ to the following system
\begin{equation}
\label{Eq_sys}
X=BZ
\end{equation}
is unique and block diagonal.
\end{theorem}
\textit{Proof}. We only need to prove that, for any data point $y\in\mathcal{S}_i$, $y\neq0$, there exists a unique $z$, $y=Bz$, where $z=[z_1^T,\cdots,z_k^T]^T$, with $z_i\neq0$ and $z_j=0$ for all $j\neq i$. Since the subspaces are independent, there exists a unique decomposition of $y$:
\begin{align*}
y   & =  0 + \cdots + y + \cdots + 0   \\
	& =  B_1z_1+\cdots+B_iz_i+\cdots +B_kz_k,
\end{align*}
where $B_iz_i\in\mathcal{S}_i$, $i=1,\cdots,k$. Thus, $B_iz_i=y$ and $B_jz_j=0$ for all $j\neq i$. Considering $B_i$ is a basis of $\mathcal{S}_i$, we have $z_i\neq 0$, $z_i$ is unique, and $z_j=0$ for all $j\neq i$.

From Theorem \ref{Thm_Blocksys}, we can learn a basis of $X$ and use it as the dictionary. If the subspaces are independent, it is very easy to get a true segmentation by solving problem (\ref{Eq_sys}). But this model cannot be directly extended for handling the data with noise which destroys the subspace structures. We consider a similar but more general model as the existing methods as follow:
\begin{equation}
\label{Eq_gen}
\min f(Z)   \  \ \text{s.t.} \  Z\in\Omega=\{Z|X=XZ\},
\end{equation}
where $f(Z)$ is a matrix function. We show that the solution to problem (\ref{Eq_gen}) is block diagonal if $f(Z)$ satisfies the following \textit{Enforced Block Diagonal (EBD)} conditions:\\ \\
\textbf{Enforced Block Diagonal Conditions} A function $f$ is defined on $\Omega(\neq\varnothing)$ which is a set of matrices. For any  $Z=\begin{bmatrix}A & B \\C & D\end{bmatrix}\in\Omega$, $Z\neq0$, where $A$ and $D$ are square matrices, $B$ and $C$ are of compatible dimension, $A$, $D\in\Omega$. Let $Z^D=\begin{bmatrix}A & 0 \\0 & D\end{bmatrix}\in\Omega$. We require
\begin{enumerate}
\renewcommand{\labelenumi}{(\theenumi)}
\item $f(Z)=f(ZP)$, for any permutation matrix $P$, $ZP\in\Omega$.
\item $f(Z)\geq f(Z^D)$, where the equality holds if and only if $B=C=0$ (or $Z=Z^D$).
\item $f(Z^D)=f(A)+f(D)$.
\end{enumerate}
\begin{theorem}
\label{Thm_Blockmin}
Assume the data sampling is sufficient \footnote{The data sampling is sufficient which makes the problem (\ref{Eq_gen}) have a nontrivial solution.}, and the subspaces are independent. If $f$ satisfies the EBD conditions (1)(2), the optimal solution(s) $Z^*$ to problem (\ref{Eq_gen}) is block diagonal:
\begin{equation*}
Z^*=\begin{bmatrix}Z_1^* & 0 & \cdots & 0 \\0 & Z_2^* & \cdots & 0 \\ \vdots & \vdots & \ddots & \vdots \\0 & 0 & \cdots & Z_k^* \end{bmatrix}
\end{equation*}
with $Z_i^*\in\mathbb{R}^{n_i\times n_i}$ corresponding to $X_i$, for each $i$. Furthermore, if $f$ satisfies the EBD conditions (1)(2)(3), for each $i$, $Z_i^*$ is also the optimal solution to the following problem:
\begin{equation}
\label{Eq_sub}
\min f(Y) \ \  \text{s.t.} \  X_i=X_iY
\end{equation}
\end{theorem}
\textit{Proof}. First, we still assume the columns of $X$ are in general position: $X=[X_1,\cdots,X_k]$, since $f(Z)=f(ZP)$, for any permutation $P$, the objective function is invariant to any permutation. Let $Z^*$ be an optimal solution to problem (\ref{Eq_gen}), we decompose $Z^*$ to two parts $Z^*=Z^D+Z^C$, where
\begin{equation*}
Z^D=\begin{bmatrix}Z_1^* & 0 & \cdots & 0 \\0 & Z_2^* & \cdots & 0 \\ \vdots & \vdots & \ddots & \vdots \\0 & 0 & \cdots & Z_k^* \end{bmatrix}, \ Z^C=\begin{bmatrix} 0 & * & \cdots & * \\ * & 0 & \cdots & * \\ \vdots & \vdots & \ddots & \vdots \\ * & * & \cdots & 0\end{bmatrix},
\end{equation*}
with $Z_i^*\in\mathbb{R}^{n_i\times n_i}$. Denote $[M]_j$ as the $j$-th column of matrix $M$. Assume $[X]_j=[XZ^*]_j\in\mathcal{S}_l$, thus $[XZ^D]_j\in\mathcal{S}_l$, $[XZ^C]_j\in\oplus_{i\neq l}\mathcal{S}_i$. But $[XZ^C]_j=[XZ^*]_j-[XZ^D]_j\in\mathcal{S}_l$, since the subspaces are independent, $\mathcal{S}_l\cap\oplus_{i\neq l}\mathcal{S}_i=\{0\}$, so $[XZ^C]_j=0$. Thus $XZ^C=0$, $XZ^D=X$, $Z^D$ is feasible for problem (\ref{Eq_gen}). By the EBD conditions (2), we have $f(Z^*)\geq f(Z^D)$. Notice $Z^*$ is optimal, $f(Z^*)\leq f(Z^D)$. Therefore, $f(Z^*)=f(Z^D)$, the equality holds if and only if $Z^*=Z^D$. Hence, $Z^*$ is block diagonal.

If the EBD conditions (3) is further satisfied, we have $f(Z^*)=\sum_{i=1}^kf(Z_i^*)$, $X=XZ^*=[X_1Z_1^*,\cdots,X_kZ_k^*]$, $X_i=X_iZ_i^*$. Hence, $Z_i^*$ is also the minimizer to problem (\ref{Eq_sub}).

From Theorem \ref{Thm_Blockmin}, it is easy to confirm SSC, LRR, MSR and SSQP are all special cases (see Table 1) by the following propositions:
\newtheorem{Proposition}{Proposition}
\begin{Proposition}
 If $f$ satisfies the EBD conditions (1)(2)(3) on $\Omega$, then also on $\Omega_1\subset\Omega$, $\Omega_1\neq\varnothing$.
\end{Proposition}
\begin{Proposition}
$\{f_i\}_{i=1}^m$ are a series of functions. For each $i$, if $f_i$ satisfies the EBD conditions (1)(2)(3) on $\Omega_i$, then also $\sum_{i=1}^m\lambda_if_i$, ($\lambda_i>0$) on $\cap_{i=1}^m\Omega_i(\neq\varnothing)$.
\end{Proposition}

\begin{table}[!t]
\caption{Criteria which satisfy the EBD conditions (1)(2)(3).}
\label{CriteriaEBD}
\centering
\begin{tabular}{c|c|c}
\hline
& $f(Z)$ & $\Omega$ \\ \hline
SSC & $||Z||_0$ or $||Z||_1$ & $\{Z|X=XZ,\text{ diag}(Z)=0\}$ \\ \hline
LRR & $||Z||_*$ & $\{Z|X=XZ\}$ \\ \hline
SSQP & $||Z^TZ||_1$ & $\{Z|X=XZ,Z\geq0,\text{diag}(Z)=0\}$ \\ \hline
MSR & $||Z||_1+\delta||Z||_*$ & $\{Z|X=XZ,\text{diag}(Z)=0\}$ \\ \hline
\multirow{2}*{Other choices}   & $(\sum_{i=1}^n\sum_{j=1}^n\lambda_{ij}|Z_{ij}|^{p_{ij}})^s$  &
\multirow{2}*{$\{Z|X=XZ,\text{diag}(Z)=0\}$} \\
	&	$\lambda_{ij}>0$, $p_{ij}>0$, $s>0$	&	\\ \hline
\end{tabular}
\end{table}

The independence of subspaces guarantees the separability of data. Theorem \ref{Thm_Blockmin} shows that many criteria can utilize such separability for correct segmentation. The minimization problem (\ref{Eq_sub}) helps us understand the within-cluster affinities which are not studied before. SSC not only enforces sparsity between-cluster but also within-cluster, which may lead to a “too sparse” solution. LRR encourages the within-cluster affinities to be of low nuclear norm (also low rank). But the physical meaning is unclear by a low rank graph.

For SSC, both $||Z||_0$ and $||Z||_1$ satisfy the EBD conditions (1)(2)(3). $||Z||_1$ is not only a good surrogate of $||Z||_0$ for optimization, but also an independent method for subspace segmentation. For the rank criterion, $\text{rank}(Z)$ does not satisfy the EBD conditions (2), which is the key to enforce sparsity between-cluster. This is also the major difference between $\text{rank}(Z)$  and $||Z||_*$. Thus, the original low rank representation (\ref{Eq_LRRrank}) may be far from enough for subspace segmentation. $||Z||_*$ has its independent ability for modeling the data from different subspaces, it is not a simple surrogate of $\text{rank}(Z)$  for subspace segmentation.

Notice SSQP requires the subspaces to be orthogonal for correct segmentation, but we show the independent subspaces assumption is enough. But the data sampling can be insufficient if the subspaces are orthogonal:
\begin{theorem}
\label{Thm_Blockorth}
If the subspaces are orthogonal, and $f$ satisfies the EBD conditions (1)(2), the optimal solution(s) to the following problem:
\begin{equation}
\label{Eq_generr}
\min ||X-XZ||_{2,p}+\lambda f(Z)
\end{equation}
must be block diagonal, where $||\cdot||_{2,p}$ is defined as $||M||_{2,p}=(\sum_j(\sum_{i=1}^nM_{ij}^2)^{\frac{p}{2}})^{\frac{1}{p}}$, $p>0$, and $\lambda>0$ is a parameter which balances the effects of two terms.
\end{theorem}
Notice the error term $||X-XZ||_{2,p}$ is not caused by noise, but the limited representational capability with insufficient data. We omit the proof here since it is very similar to Theorem 2 in \cite{SSQP}, one only needs to replace $||Z^TZ||_1$ as $f(Z)$ and complete the proof similarly. Theorem \ref{Thm_Blockorth} guarantees the block diagonal structure of the solution to problem (\ref{Eq_generr}) from insufficient data. Though the orthogonal subspaces assumption is possibly violated in real data, Theorem \ref{Thm_Blockorth} provides a theoretical low bound for correct segmentation from insufficient data.

\section{Subspace Segmentation via LSR}

\subsection{Least Squares Regression}

The theoretical analysis in Section \ref{Se_TheoAna} not only summarizes some existing works, but also helps us design a new criterion. Theorem \ref{Thm_Blockmin} shows that many criteria guarantee the block diagonal property between-cluster. But the within-cluster affinities are also very important for subspace segmentation. Sparse representation encourages sparsity not only between-cluster, but also within-cluster, thus it misses the important correlation structure in the data. However, most data exhibit strong correlations. The subspace segmentation problem studied in this paper assumes data are drawn from a union of subspaces. If the sampling data are sufficient, they tend to be highly correlated. For modeling such data, we present the \textit{Least Squares Regression (LSR)} method as follow:
\begin{equation}
\label{LSR}
\min ||Z||_F \ \ \text{s.t.} \ X=XZ, \ \text{diag}(Z)=0.
\end{equation}
Here $||Z||_F$ denotes the Frobenius norm of $Z$, \emph{i.e.} $||Z||_F=(\sum_{i=1}^n\sum_{j=1}^nZ_{ij}^2)^{\frac{1}{2}}$, which  is a special case of $(\sum_{i=1}^n\sum_{j=1}^n\lambda_{ij}|Z_ij|^{p_{ij}})^s$. As corollaries to Theorem \ref{Thm_Blockmin} and Theorem \ref{Thm_Blockorth}, we have

\begin{theorem}
Assume the data sampling is sufficient, and the subspaces are independent. The optimal solution $Z^*$ to problem (\ref{LSR}) is block diagonal:
\begin{equation*}
Z^*=\begin{bmatrix}Z_1^* & 0 & \cdots & 0 \\0 & Z_2^* & \cdots & 0 \\ \vdots & \vdots & \ddots & \vdots \\0 & 0 & \cdots & Z_k^* \end{bmatrix},
\end{equation*}
where $Z_i^*\in\mathbb{R}^{n_i\times n_i}$ is also the optimal solution to the following problem:
\begin{equation}
\label{Eq_sub}
\min ||Y||_F \ \  \text{s.t.} \  X_i=X_iY, \ \text{diag}(Y)=0.
\end{equation}
\end{theorem}
\begin{theorem}
If the subspaces are orthogonal, the optimal solution to the following problem
\begin{equation}
\min ||X-XZ||_{2,p}+\lambda ||Z||_F^2
\end{equation}
is block diagonal, where $p>0$, $\lambda>0$.
\end{theorem}

The above two theorems show LSR can reveal the true subspace membership under some data assumption.

\subsection{LSR with Noise}
The data from real applications are always contaminated with noise. Similar to SSC, we use the Frobenius norm to penalize the noise as follow:
\begin{equation}
\label{LSRnoise1}
\min ||X-XZ||_F^2+\lambda||Z||_F^2 \ \  \text{s.t.} \ \text{diag}(Z)=0,
\end{equation}
where $\lambda>0$ is a parameter used to balance the effects of the two parts. Problem (\ref{LSRnoise1}) can be efficiently solved by the following analytical solution:
\begin{theorem}
\label{Thm_LSR1}
The optimal solution to problem (\ref{LSRnoise1}) is
\begin{equation}
\label{Eq_LSRsolu1}
Z^*=-D(\text{diag}(D))^{-1} \text{ and diag}(Z^*)=0,
\end{equation}
where $D=(X^TX+\lambda I)^{-1}$.
\end{theorem}

The proof of Theorem \ref{Thm_LSR1} is presented in Appendix.

The constraint diag$(Z)=0$ can be removed from problem (\ref{LSRnoise1}) which leads to another formulation of LSR as follow:
\begin{equation}
\label{Eq_LSRnoise2}
\min ||X-XZ||_F^2+\lambda||Z||_F^2.
\end{equation}
It also has an analytical solution:
\begin{equation}
\label{Eq_LSRsolu2}
Z^*=(X^TX+\lambda I)^{-1}X^TX.
\end{equation}
The problem (\ref{Eq_LSRnoise2}) is actually the well known Tikhonov regularization \cite{Tikhonov} or ridge regression \cite{ridge}. We will discuss why it is better than SSC and LRR for subspace segmentation in next subsection.
\subsection{Why LSR?}
The motivation by using LSR for subspace segmentation is that it tends to shrink coefficients of correlated data and groups them together. LSR exhibits the grouping effect that the coefficients of a group of correlated data are approximately equal. The grouping effect of LSR is stated in the following theorem:
\begin{theorem}
\label{thm_groupingeffect}
Given a data vector $y\in\mathbb{R}^d$, data points $X\in\mathbb{R}^{d\times n}$ and a parameter $\lambda$. Assume each data point of $X$ are normalized. Let $z^*$ be the optimal solution to the following LSR (in vector form) problem:
\begin{equation}
\label{Eq_LSRvector}
\min ||y-Xz||_2^2+\lambda ||z||_2^2.
\end{equation}
We have
\begin{equation}
\label{Eq_groupingeffect}
\frac{||z_i^*-z_j^*||_2}{||y||_2}\leq\frac{1}{\lambda}\sqrt{2(1-r)},
\end{equation}
where $r=x_i^Tx_j$ is the sample correlation.
\end{theorem}

The proof of Theorem \ref{thm_groupingeffect} is presented in Appendix.

The grouping effect of LSR presented in the above theorem shows that the solution is correlation dependent. If $x_i$ and $x_j$ are highly correlated, \emph{i.e.} $r=1$ (if $r=-1$ then consider $-x_j$), Theorem \ref{thm_groupingeffect} says that the difference between the coefficient paths of $x_i$ and $x_j$ is almost 0. Thus $x_i$ and $x_j$ will be grouped in the same cluster.

Sparse representation does not have the grouping effect and even it is unstable. The grouping effect of LRR is still unclear \footnote{The original LRR by rank minimization (\ref{Eq_LRRrank}) does not have the grouping effect.}, but note that (\ref{Eq_groupingeffect}) is a tight bound and so we can expect LSR to possess greater grouping capabilities. It is interesting that the solution (\ref{Eq_LSRsolu2}) of LSR is also of low rank, $\text{rank}(Z^*)=\text{rank}(X)$. But the effectiveness of LSR for subspace segmentation comes from its grouping effect, it has nothing to do with the property of its low rankness.

\subsection{Algorithm of Subspace Segmentation by LSR}
Similar to SSC and LRR, our method is also a spectral clustering-based method. We first solve the LSR problem by (\ref{Eq_LSRsolu1})  or (\ref{Eq_LSRsolu2}) (the solution is denoted by $Z^*$), then define the affinity matrix as $(|Z^*|+|(Z^*)^T|)/2$. The spectral clustering algorithm such as Normalized Cuts \cite{NCuts} is employed to produce the ultimate segmentation results. In summary, we have the entire Least Squares Regression (LSR) algorithm for segmenting data drawn from multiple subspaces in Algorithm 1.
\begin{algorithm}[t]
\label{SSviaLSR}
\caption{Subspace Segmentation via LSR}
\textbf{Input:} data matrix $X$, number of subspaces $k$.
\begin{enumerate}
\item Solve the LSR problem by (\ref{Eq_LSRsolu1}) or (\ref{Eq_LSRsolu2}).
\item Define the affinity matrix by $(|Z^*|+|(Z^*)^T|)/2$.
\item Segment the data into $k$ subspaces by Normalized Cuts.
\end{enumerate}
\end{algorithm}
\section{Experimental Verification}
\label{sec:blind}
In this section, we evaluate LSR \footnote{The Matlab code: \url{http://home.ustc.edu.cn/~canyilu/}} on the Hopkins 155 motion database and Extended Yale Database B, in comparison with SSC \footnote{The Matlab code: \url{http://www.vision.jhu.edu/code/}, it uses the CVX package for solving the sparse representation problem, we instead use the SPAMS package (\url{http://www.di.ens.fr/willow/SPAMS/index.html}), which is very efficient due to its implementation by C++. It is not fair, but SSC is still slower than LRR and LSR.} and LRR \footnote{The Matlab code: \url{http://sites.google.com/site/guangcanliu/}}. We implement two versions of LSR, LSR1 denotes the LSR algorithm by (\ref{Eq_LSRsolu1}), and LSR2 denotes the LSR based on (\ref{Eq_LSRsolu2}). All experiments are carried out by by using Matlab on a PC with 2.4GHz CPU and 2GB RAM.

\subsection{Motion Segmentation}
We apply LSR on the Hopkins 155 database \footnote{\url{http://www.vision.jhu.edu/data/hopkins155/}}. It consists of 156 sequences of two or three motions (a motion corresponding to a subspace). Each sequence is a sole segmentation task and so there are 156 subspace segmentation tasks totally. We first use PCA to project the data onto a 12-dimensional subspace. Then SSC, LRR and LSR are employed for comparison. The parameter is manually tuned for each method and we report the best result. Table \ref{Tab_Hop} shows the parameter, segmentation errors, and running time of each method on the Hopkins 155 database. Notice the running time show in the Table \ref{Tab_Hop} is the cost that we compute the affinity matrix by different methods.

\begin{table}[!t]
\centering
\caption{The parameter, segmentation errors (\%), and running time (s) of each method on the Hopkins 155 Database}
\label{Tab_Hop}
\centering
\begin{tabular}{c|c|c c c c}
\hline
\multicolumn{2}{c|}{ } & SSC & LRR & LSR1 & LSR2 \\ \hline
\multicolumn{2}{c|}{ Parameter $\lambda$ } & $2\times10^{-3}$ & 2.4 & $4.8\times10^{-3}$ & $4.6\times10^{-3}$\\ \hline
\multirow{4}*{Error} & Max		& 39.53	&	36.36	&	36.36	&	36.36 	\\ \cline{2-6}
					 & Mean 	& 4.02	&	3.23	&	\textbf{2.50}	&	2.84	\\ \cline{2-6}
					 & Median 	& 0.90	&	0.50	&	\textbf{0.31}	&	0.34	\\ \cline{2-6}
					 & STD 		& 10.04	&	6.06	&	\textbf{5.62}	&	6.16	\\ \hline
\multicolumn{2}{c|}{ Running times } & 149.70 &	129.30	&	24.33	&	\textbf{21.35}	\\ \hline
\end{tabular}
\end{table}

\subsection{Face Clustering}
We evaluate LSR on the Extended Yale Database B [25]. In this experiment, we use the first 5 and 10 classes data, each class contains 64 images. The images are resized into $32\times32$. Then the data are projected onto a $5\times6$-dimensional subspace for 5 classes clustering problem by PCA, and a $10\times6$-dimensional subspace for 10 classes clustering problem. The three methods all perform well on the above setting. Table 3 lists the parameter, segmentation accuracies, and running time of each method on the Extended Yale Database B.

\begin{table}[!t]
\centering
\caption{The parameter, segmentation accuracies (\%), and running time (s) of each method on the Extended Yale Database B}
\label{Tab_YaleB}
\centering
\begin{tabular}{c|c|c c c c}
\hline
\multicolumn{2}{c|}{ } & SSC & LRR & LSR1 & LSR2 \\ \hline
\multirow{2}*{ Parameter $\lambda$ } & 5	&	0.05	&	0.15	&	0.4		&	0.4 \\ \cline{2-6}
									 & 10	&	0.05	&	1.5		&	0.004	&	0.004	\\	\hline
\multirow{2}*{ Accuracy }			 & 5	&	76.88	&	81.88	&	88.13	&	\textbf{91.56}	\\	\cline{2-6}
									 & 10	&	47.81	&	65.00	&	70.16	&	\textbf{72.34}	\\	\hline
\multirow{1}*{ Running }			 & 5	&	0.657	&	0.602	&	0.018	&	\textbf{0.009}	\\	\cline{2-6}
\multirow{1}*{ time }			 & 10	&	4.760	&	2.261	&	0.101	&	\textbf{0.045}	\\	\hline
\end{tabular}
\end{table}

\subsection{Experimental Results and Discussions}

Based on the experimental results shown in Table \ref{Tab_Hop} and Table \ref{Tab_YaleB}, we have the following observations and discussions:
\begin{enumerate}
\renewcommand{\labelenumi}{(\theenumi)}
\item LSR outperforms SSC and LRR on the Hopkins 155 database and Extended Yale Database B. The advantage of LSR is mainly due to its grouping effect for modeling the correlation structure of data. Notice the motion data exhibit strong correlations, the dimension of each affine subspace is at most three \cite{SSC}. The correlation structure of face images has been widely used for face recognition \cite{YaleB} \cite{SRC}.

\item LSR is robust. Theorem 1.9 in \cite{OptML} shows that problem (\ref{Eq_LSRnoise2}) is equivalent to a robust optimization problem, when the data $X$ is subjected to a bounded matrix disturbance with a Frobenius norm.

\item Beyond the performance and robustness, the experimental results show that LSR is more efficient than SSC and LRR. SSC solves the $\ell^1$-minimization problem, which is non-smooth and requires much computational cost. The optimization of LRR by inexact ALM which demands hundreds of singular value decomposition and the convergence is not strictly proved in theory.
\item The most important is that LSR is simpler and better. The previous models for subspace segmentation, SSC, LRR, MSR and SSQP, are unnecessary sophistication.
\end{enumerate}

\section{Conclusions}

This paper explores the subspace segmentation problem. We first theoretically provide an Enforced Block Diagonal (EBD) conditions, and show that if the subspaces are independent, many criteria which satisfy the EBD conditions always produce a block diagonal solution. The EBD conditions are general and easy to be satisfied, the existing methods, SSC, LRR, MSR and SSQP are all special cases. Furthermore, the data sampling can be insufficient when the subspaces are orthogonal. Second, considering that sufficient data which can be characterized by linear subspace are usually highly correlated, we further propose the Least Squares Regression (LSR) method which takes advantage of the correlation of data. We theoretically show that the grouping effect of LSR makes it group the highly correlated data together, and also it is robust to noise. Experimental results on real data demonstrate that LSR is efficient and effective, with comparison to the state-of-the-art subspace segmentation methods SSC and LRR. LSR is simple while the existing methods are sophisticated.

\section*{Acknowledgment} This research is partially supported by the Singapore National Research Foundation under its International Research Centre @ Singapore Funding Initiative and administered by the IDM Programme Office, the grants of the NSFC, Nos. 61005010, 61005007, 31071168, 61133010 and 61100161.

\section*{Appendix}
\subsection*{Proof of Theorem \ref{Thm_LSR1}}
\textit{Proof.} Assume $X=[x_1,x_2,\cdots,x_n]\in\mathbb{R}^{d\times n}$. Remove the $i$-th column of $X$, we get $Y_i=[x_1,\cdots,x_{i-1},x_{i+1},\cdots,x_n]$. The $i$-th column of $Z^*$ is $[Z^*]_i=E_iY_i^Tx_i$, where $E_i=(Y_i^TY_i+\lambda I)^{-1}$. But it is not efficient to compute an inverse of matrix for each $x_i$. We will show how to obtain $[Z^*]_i$ from $D=(X^TX+\lambda I)^{-1}$ which can be pre-calculated. First, we arrange $X$ as $XP=[Y_i \ x_i]$, where $P$ is a permutation matrix, $PP^T=P^TP=I$. Thus we have
\begin{equation}
\label{Eq_App1}
[P^T(X^TX+\lambda I)P]^{-1}=P^TDP.
\end{equation}
On the other hand, we can compute $[P(X^TX+\lambda I)P]^{-1}$ by using the Woodbury formula \cite{MatCom} as follow:
\begin{align*}
  & [P^T(X^TX+\lambda I)P]^{-1} \\
= & \begin{bmatrix} Y_i^TY_i+\lambda I & Y_i^Tx_i \\ x_i^TY_i & x_i^Tx_i+\lambda \end{bmatrix}^{-1} \\
= & \begin{bmatrix} E_i & 0 \\ 0 & 0 \end{bmatrix} + \beta_i \begin{bmatrix} b_ib_i^T & b_i \\ b_i^T & 1 \end{bmatrix}
\end{align*}
where
\begin{equation*}
b_i=-[Z^*]_i,
\end{equation*}
\begin{equation*}
\beta_i=x_i^Tx_i+\lambda-x_i^TY_iE_iY_i^Tx_i.
\end{equation*}
Thus $[Z^*]_i=-b_i$ is what we need, and we can get $b_i$ from (\ref{Eq_App1}) by considering the property of $P$, we get
\begin{equation*}
Z_{ji}^*=
\begin{cases}
-\frac{D_{ji}}{D_{ii}}, \quad & j\neq i, \\
0, \quad	 & j=i.
\end{cases}
\end{equation*}
The solution can be rewritten as $Z^*=-D(\text{diag}(D))^{-1}$ , $\text{diag}(Z^*)=0$.
\subsection*{Proof of Theorem \ref{thm_groupingeffect}}
\textit{Proof.} Let $L(z)=||y-Xz||_2^2+\lambda||z||_2^2$. Since $z^*$ is the optimal solution to problem (\ref{Eq_LSRvector}), it satisfies
\begin{equation}
\left. \frac{\partial L(z)}{\partial z_k} \right| _{z=z^*}=0.
\end{equation}
Thus we have
\begin{equation}
\label{Eq_App2}
-2x_i^T(y-Xz^*)+2\lambda z_i^*=0,
\end{equation}
\begin{equation}
\label{Eq_App3}
-2x_j^T(y-Xz^*)+2\lambda z_j^*=0.
\end{equation}
Equation (\ref{Eq_App2}) and (\ref{Eq_App3}) gives
\begin{equation}
\label{Eq_App4}
z_i^*-z_j^*=\frac{1}{\lambda}(x_i^T-x_j^T)(y-Xz^*).
\end{equation}
Since each column of $X$ is normalized, $||x_i-x_j||_2=\sqrt{2(1-r)}$ where $r=x_i^Tx_j$. Notice $z^*$ is optimal to problem (\ref{Eq_LSRvector}), we get
\begin{equation}
||y-Xz^*||_2^2+\lambda||z^*||_2^2=L(z^*)\leq L(0)=||y||_2^2
\end{equation}
Thus $||y-Xz^*||_2\leq ||y||_2$. Then equation (\ref{Eq_App4}) implies
\begin{equation}
\frac{||z_i^*-z_j^*||_2}{||y||_2}\leq\frac{1}{\lambda}\sqrt{2(1-r)}.
\end{equation}

\bibliographystyle{splncs}
\bibliography{LSR}

%{\footnotesize
%\bibliographystyle{splncs}
%\bibliography{nncw_image_cvpr}
%}

\end{document}